\documentclass{article}
\PassOptionsToPackage{table,xcdraw}{xcolor}
\usepackage{xcolor}
\usepackage{bbm}
\usepackage{multirow}
\usepackage[english]{babel}

\usepackage[letterpaper,top=2cm,bottom=2cm,left=3cm,right=3cm,marginparwidth=1.75cm]{geometry}

\usepackage{amsmath}
\usepackage{graphicx}
\usepackage[colorlinks=true, allcolors=blue]{hyperref}
\usepackage{booktabs} % To thicken table lines
\usepackage{caption}
\newcommand{\deluxesystem}{\textsc{LLMSelector}}

\newcommand{\eat}[1]{}

\usepackage{graphicx}
\usepackage{amssymb}
\usepackage{multirow}
\usepackage{bigstrut,morefloats}

\usepackage[resetlabels,labeled]{multibib}
\usepackage{caption}
\usepackage{subcaption}

\usepackage[most]{tcolorbox}
\newcommand{\E}{\mathbb{E}}

\usepackage{amsmath}
\usepackage{amsthm}
\usepackage{thmtools}

\theoremstyle{definition}

\usepackage[boxruled,linesnumbered]{algorithm2e}

\usepackage{amsthm}

\usepackage[most]{tcolorbox} % loads the tcolorbox package with most common features

\newtcolorbox[auto counter, number within=section]{mybox}[2][]{
    colback=blue!5!white,    % background color
    colframe=blue!75!black,  % frame color
    fonttitle=\bfseries,     % title font
    title=Main Contributions, % box title
    #1                      % allow for additional options
}

\newtcolorbox[auto counter, number within=section]{LLMdiagnoser}[2][]{
    colback=blue!5!white,    % background color
    colframe=blue!75!black,  % frame color
    fonttitle=\bfseries,     % title font
    title=LLM diagnoser prompt, % box title
    #1                      % allow for additional options
}

\usepackage{multirow}
\usepackage{array}
\usepackage{booktabs} % For professional looking tables

\usepackage{graphicx}
\usepackage[authoryear]{natbib}

\usepackage{amsmath}

\newcommand{\question}{q}
\newcommand{\answer}{a}
\newcommand{\query}{z}

\newcommand{\Dataset}{\mathcal{D}}

\newcommand{\Datasettrain}{\mathcal{D}_{Tr}}

\newcommand{\allocation}{f }

\newcommand{\allocationspace}{F }

\newcommand{\Performance}{\mathit{P}}

\newcommand{\performance}{\mathit{p}}

\newcommand{\decompose}{\mathit{h}}

\usepackage{pifont}
\theoremstyle{plain}
\newtheorem{theorem}{Theorem}[section]

\theoremstyle{definition}

\theoremstyle{remark}

\title{Optimizing Model Selection for Compound AI Systems}
\author{Lingjiao Chen$^{\dagger,\circ}$, Jared Quincy Davis$^{\circ}$, Boris Hanin$^{\S}$\\\\ Peter Bailis$^\ddagger$, Matei Zaharia$^\ddagger$, James Zou$^{\circ}$, Ion Stoica$^\ddagger$\\\\
$^\dagger$Microsoft Research, $^\circ$Stanford University,\\ $^\S$Princeton University, $^\ddagger$University of California, Berkeley}
\date{}

\renewcommand{\cite}[1]{\citep{#1}}

\begin{document}
\maketitle

\begin{abstract}
Compound AI systems that combine multiple LLM calls, such as self-refine and multi-agent-debate, achieve strong performance on many AI tasks. We address a core question in optimizing compound systems: for each LLM call or module in the system, how should one decide which LLM to use? We show that these LLM choices have a large effect on quality, but the search space is exponential. We propose \deluxesystem{}, an efficient framework for model selection in compound systems, which leverages two key empirical insights: (i) end-to-end performance is often monotonic in how well each module performs, with all other modules held fixed, and (ii) per-module performance can be estimated accurately by an LLM. Building upon these insights, \deluxesystem{} iteratively selects one module and allocates to it the model with the highest module-wise performance, as estimated by an LLM, until no further gain is possible. \deluxesystem{} is applicable to any compound system with a bounded number of modules, and its number of API calls scales linearly with the number of modules, achieving high-quality model allocation both empirically and theoretically. Experiments with popular compound systems such as multi-agent debate and self-refine using LLMs such as GPT-4o, Claude 3.5 Sonnet and Gemini 1.5  show that \deluxesystem{} confers 5\%-70\% accuracy gains compared to using the same LLM for all modules.
 
\end{abstract}

\section{Introduction}
\begin{figure*}[!ht]
    \centering
    \includegraphics[width=0.99\linewidth]{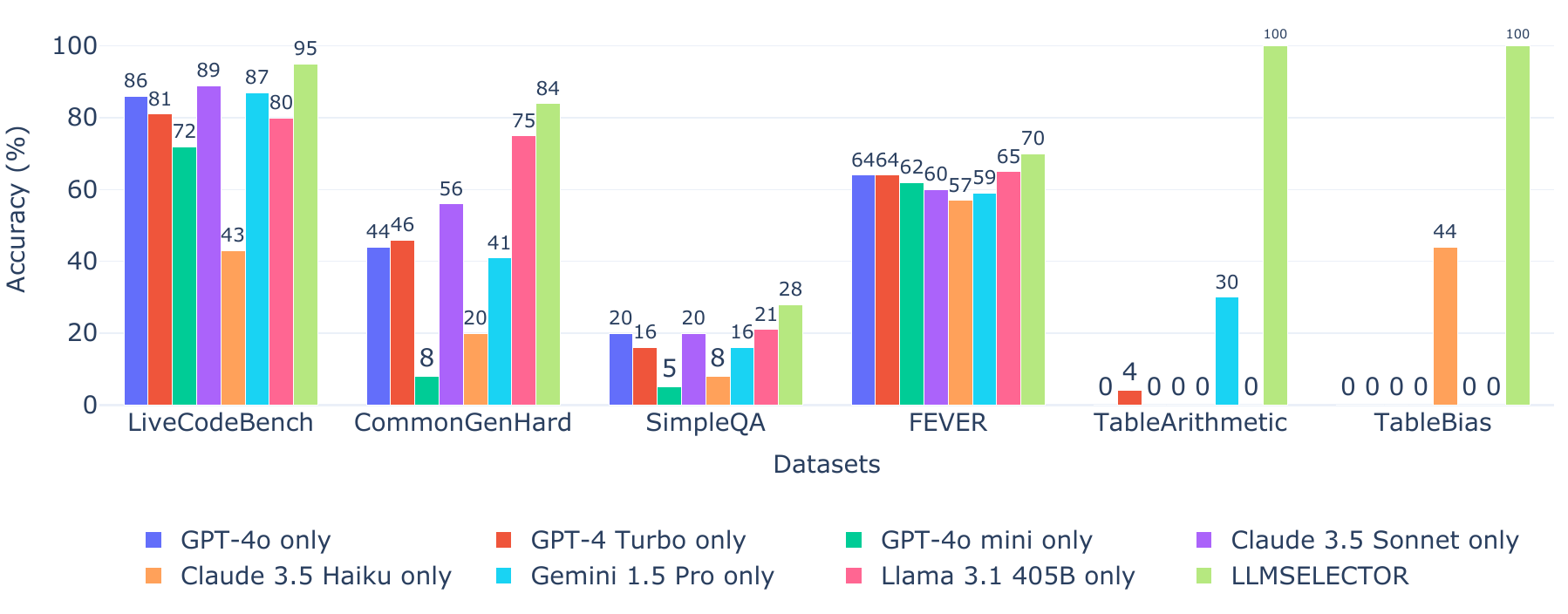}
    \caption{\deluxesystem{} outperforms compound AI systems that always call the same LLM. Here we study three compound systems, namely, self-refine (on LiveCodeBench and GCH), multi-agent-debate (on SimpleQA and FEVER), and locate-solve (on TableArithmetic and TableBias).  \deluxesystem{} achieves 5\%-70\% accuracy gains over allocating any model alone by allocating different models to different modules in these compound systems.
    }
    \label{fig:deluxeagent:intro}
\end{figure*}
Researchers and developers are increasingly leveraging large language models (LLMs) by composing multiple LLM calls in a compound AI system to tackle complex tasks~\cite{multi_agent_debate_2024,zhang2024chain,madaan2024self,deepmind2025alphacode2,shinn2024reflexion,renze2024self,compound-ai-blog}. For example, a common practice is to use one LLM call to generate one initial answer, one LLM call to give feedback, and one more call to refine the answer based on the feedback, known as self-refine~\cite{renze2024self,madaan2024self,ji2023towards}.
Another example is multi-agent debate~\cite{multi_agent_debate_2024,liang-etal-2024-encouraging,khan2024debating}, where multiple LLM calls are made to propose initial answers and then debate which ones are correct.  Compared to monolithic models, significant improvements are possible because the compound systems decompose challenging tasks into simpler sub-tasks, and perform one LLM call for each sub-task.

\eat{Despite the commonality of invoking multiple LLM calls, there is little understanding of \textit{model selection} for compound AI systems. In particular, existing work often focuses on optimizing prompts used in individual modules and/or module interactions, and uses the same LLM for all modules~\cite{}. While simplifying the design, overlooking model selection leaves many important questions open. For example, does allocating different models to different modules improve a compound system's performance?  If so, why and by how much? Given a pool of LLMs, how can we find such a model allocation without an exhaustive search?  These questions indicate a significant gap in understanding and optimizing compound AI systems.

As a first step towards fulfilling this gap, we systematically study model selection focusing on static compound AI systems, i.e., those where the number of modules is fixed. We have found module-wise heterogeneity: an LLM A might be better than an LLM B for module 1, but worse for module 2. This suggests that generation improvement is possible by allocating different models to different modules. Consider, for example, the self-refine system as shown in Figure \ref{fig:deluxeagent:intro}. Allocating GPT-4o to the generator and refiner modules and Claude 3.5 Sonnet to the critic module leads to better performance than always using GPT-4o or Claude 3.5 Sonnet alone.
This is because of the module-wise heterogeneity:  GPT-4o is better at answering and refining, while Claude 3.5 Sonnet is better at providing feedback.
}

Most existing work on improving compound systems focuses on optimizing prompts used in individual modules and/or module interactions, while using the same LLM for all modules~\cite{khattab2024dspy,yuksekgonul2024textgrad,wu2023autogen}. %For example, DSPy~\cite{khattab2024dspy} takes a compound AI system's architecture and one LLM as input, and jointly optimizes all modules' prompts via Bayesian optimization.
While this simplifies compound system design, it also leaves several important questions unaddressed.
Does using different models across modules 
improve a compound system's performance? If so, why and by how much? 
Given a pool of LLMs, can we find the best model each module should use without exhaustive search?

As a first step towards answering such questions, we systematically study model selection in static compound AI systems, i.e., those where the number of modules, the sequencing of module calls, and the mapping between modules and models are fixed.
In this context, we indeed find that allocating different LLMs to different modules leads to substantially higher performance than allocating the same LLM to all modules. 
As an example, consider again the self-refine system~\cite{madaan2024self} consisting of three modules: a generator, a critic, and a refiner. LLM A may be better at providing feedback but worse at generating and refining answers than LLM B. In this case, allocating LLM A for the critic and LLM B for the generator and refiner is better than allocating either one to all modules.

Then we formulate the model selection problem (MSP), i.e., identifying the best model each module should use to maximize the overall performance. MSP is challenging in principle, as it is infeasible to exhaustively search the exponentially large space of all model choices. Our insights are that, in many cases, (i) the end-to-end performance can be monotonic in per-module performance, and (ii) per-module performance can be estimated accurately by an LLM diagnoser. This motivates us to design \deluxesystem{}, a principled framework that optimizes MSP for any static compound AI systems given a training budget. \deluxesystem{} iteratively nominates one module and allocates to it the model with the best module-wise performance, as estimated by an LLM diagnoser. One benefit is that \deluxesystem{} is applicable to any compound AI system whose number of modules is fixed. Furthermore, \deluxesystem{} only incurs a manageable amount of LLM calls. In fact, we provide mathematical conditions under which \deluxesystem{} finds the optimal solution to MSP with the number of LLM calls linear to the number of modules (Section \ref{sec:deluxeagent:Method}). 

We conduct systematic experiments on a diverse set of compound AI systems using real-world LLM APIs including GPT-4o, Claude 3.5 Sonnet, and Gemini 1.5 Pro. Perhaps surprisingly, we have found that different model choices have a significant effect on compound systems' performance. In fact, \deluxesystem{} offers 5\%-70\% performance gains compared to allocating the same LLM to all modules (Figure \ref{fig:deluxeagent:intro}). While not optimizing prompts, \deluxesystem{} also outperforms advanced techniques specializing in prompt optimization (Table \ref{tab:deluxeagent:mainresult} in Section \ref{sec:deluxeagent:Exp}). This further highlights the importance of model selection for compound AI systems. 

In short, our main contributions are:
\begin{itemize}
\item \textbf{Model selection problem.} We formulate the model selection problem (MSP) for compound AI systems, an increasingly important but under-explored problem. \\

\item \textbf{The \deluxesystem{} framework.} To optimize MSP, we propose \deluxesystem{}, a principled framework that iteratively chooses one module and allocates to it the model with the highest module-wise performance estimated by an LLM. \\

\item \textbf{Model choices matter.} Through extensive experiments on practical compound systems using real-world LLM APIs including GPT-4o, Claude 3.5 Sonnet, and Gemini 1.5 Pro, we have found that choosing different models can substantially affect (up to 100\%) a compound AI system's performance.

\item \textbf{\deluxesystem{} finds excellent choices.} Systematical experiments have shown that \deluxesystem{} identifies model choices that outperform allocating the same LLM to all modules by 5\%-70\%.\\

\item \textbf{Open-source artifacts.} We release\footnote{\url{https://github.com/LLMSELECTOR/LLMSELECTOR}} our code and data, including compound systems' intermediate outputs generated by commercial LLM APIs.
\end{itemize}

\section{Related Work}\label{sec:deluxeagent:Relatedwork}

\paragraph{Compound AI system optimization.} Prompt engineering and module interaction design is a central topic of compound AI system optimization. While existing work often relies on manually tuning them~\cite{deepmind2025alphacode2,shinn2024reflexion,zhou2024agents2,pryzant2023automatic,fourney2024magentic,zhao2024expel,lu2024chameleon,zhao2024expel}, recent work studies how to automate this process, such as DSPy~\cite{khattab2024dspy}, Textgrad~\cite{yuksekgonul2024textgrad}, and Autogen~\cite{wu2023autogen}. For example, DSPy uses Bayesian optimization to adjust prompts for all modules, while Textgrad uses textual feedback to optimize prompts for individual modules. On the other hand, our work focuses on model selection, a third axis for compound system optimization, complementary to prompt optimization and module interaction design. 

\paragraph{Model market utilization.} Model market utilization studies how to use all available (proprietary and open-source) models for downstream tasks~\cite{lu2024merge,ramirez2024optimising,miao2023towards}. Extensive work has built various techniques to utilize different models, such as model cascade~\cite{chen2023frugalgpt}, model routing~\cite{hu2024routerbench,stripelis2024tensoropera}, and mixture-of-experts~\cite{wang2024mixture}. While they mainly focus on  \textit{single-stage} tasks such as classification~\cite{chen2020frugalml,huang2025thriftllm} and question answering~\cite{chen2023frugalgpt,shekhar2024towards}, we study model utilization for compound AI systems requiring \textit{multiple stages}. This is a much more challenging problem as the search space is much larger. 

\paragraph{Model selection.} Model selection is a critical part of classic ML and has been extensively studied in the literature~\cite{kohavi1995study,akaike1974new,elsken2019neural}. %It involves identifying the most suitable model from a set of candidates via on performance metrics, generalization ability, and computational efficiency. %Cross-validation and bootstrap~\cite{kohavi1995study}, Akaike information criterion~\cite{akaike1974new}, neural architecture search~\cite{elsken2019neural} and many other techniques have been developed during the past few decades. 
While classic techniques focus on model selection for one ML task, compound systems involve multiple ML tasks. Thus, model selection becomes more challenging as the search space is exponentially large in the number of tasks.  

\paragraph{LLM-as-a-judge.} LLMs have been increasingly used for evaluating and judging complex generations, a phenomenon termed LLM-as-a-judge. Researchers have extensively studied how LLM judges align with human preferences in real-world scenarios~\cite{zheng2023judging,shankar2024validates}, how to improve its quality~\cite{kim2023prometheus}, how to evaluate it~\cite{chiang2024chatbot,chen2024mllm,zeng2023evaluating}, as well as many other applications~\cite{johri2025evaluation,dhole2024conqret,gu2024survey,zhou2024llm}. In this paper, we find a novel use case of LLM-as-a-judge: diagnosing module-wise performance to accelerate the model allocation search process. 

\section{Compound AI Systems}\label{sec:deluxeagent:Prelim}
\begin{figure}[ht]
    \centering
    \includegraphics[width=0.99\linewidth]{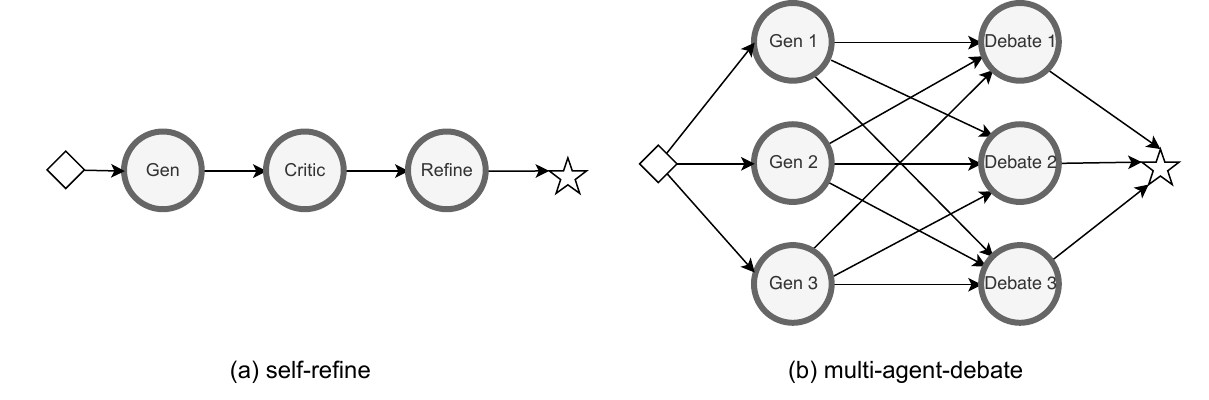}
    \caption{Examples of static compound AI systems. (a) self-refine system. (b)  multi-agent-debate system. The diamond and star represent the input and output modules, and the circles represent the LLM modules. Directed lines represent data flow, and we omit most query inputs for simplicity.}
    \label{fig:deluxeagent:demo}
\end{figure}
\paragraph{Static Compound AI systems.} As defined by ~\cite{compound-ai-blog}, compound AI systems address AI tasks by synthesizing multiple components that interact with each other. Here, we denote a static compound AI system by a directed acyclic graph $G\triangleq (V,E)$, where each node $v\in V$ denotes one module, and each directed edge $e\triangleq (u,v)\in E$ indicates that the output from module $u$ is sent to module $v$ as input. Without loss of generality, we assume a final output module that generates the final output without any output edges, and an input module representing the input query which receives no input edges. 

\paragraph{LLM modules.} An LLM module is a module that utilizes an LLM to process the inputs. It typically concatenates all inputs as a text snippet (via some prompt template), obtain an LLM's response to this snippet, and send the response as output (potentially after some postprocessing). Throughout this paper, all modules are LLM modules to simplify notations. In practice, if a  module is not an LLM module, one can either merge it into an LLM module (e.g., a module that postprocesses output from some LLM module), or convert it into an LLM module by conceptually ``adding'' an LLM to the module.

\paragraph{Examples.} Consider two examples of static compound AI systems, self-refine and multi-agent-debate. Self-refine, as shown in Figure \ref{fig:deluxeagent:demo}(a), consists of three modules: a generator, a critic, and a refiner. Given a query, the generator produces an initial answer. The critic provides feedback on the initial answer, and the refiner uses the feedback to improve the initial answer. Figure \ref{fig:deluxeagent:demo}(b) shows the architecture of a six-module system: multi-agent-debate. Here, three generators first give their initial answers to a question, then three debaters debate with each other based on these initial answers. Refinements and debates can be iterative, but we focus on only one iteration for simplicity.  

\paragraph{Notations.} Table \ref{tab:deluxeagent:notation} list our notations. We also use $f_{i\rightarrow k}$ to indicate a function that is the same as function $f$ except that the value $i$ is mapped to the value $k$.

\begin{table}[ht!]
\centering
\caption{Notations. 
}
\begin{tabular}{@{}ll@{}}
\toprule
\textbf{Symbol} & \textbf{Description} \\ \midrule
$G = (V, E)$   & A compound AI system \\ \midrule
$|V|$            & Number of LLM modules \\ \midrule
$M$            & The set of LLMs \\ \midrule
$\allocation:V\mapsto M$          & A model allocation  \\ \midrule
$\query$          & One task \\ \midrule
$\Performance(f)$            & End-to-end performance \\ \midrule
$\performance(f,z)$            & End-to-end performance on $\query$\\ \midrule
$p_i(f, \query)$    & $i$th module's performance on $\query$ \\
\bottomrule
\end{tabular}
\label{tab:deluxeagent:notation}
\end{table}
\section{Modeling and Optimizing Model Selection}\label{sec:deluxeagent:Method}

\begin{figure*}[!ht]
    \centering
    \includegraphics[width=0.99\linewidth]{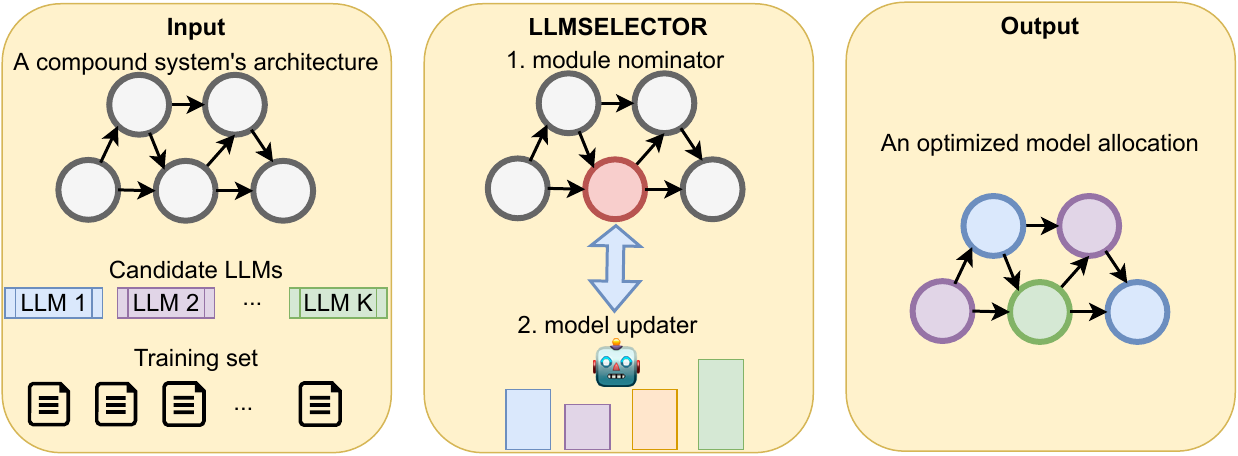}
    \caption{\deluxesystem{} Workflow. \deluxesystem{} takes as input a compound AI system, a pool of candidate LLMs, a training dataset consisting of question-answer pairs, and a training budget. Then \deluxesystem{} iteratively nominates one module and allocates to it the model with the highest module-wise performance estimated by an LLM. This is repeated until the budget is reached or no performance gain is possible. Finally, \deluxesystem{} returns an optimized model allocation. }
    \label{fig:deluxeagent:method}
\end{figure*}
This section presents how to model and optimize model selection for static compound AI systems. 

\subsection{Problem Statement}
Consider a static compound AI system $G=(V,E)$ and a set of LLMs  $M\triangleq \{1,2,\cdots, |M|\}$ to use. Let $\allocationspace:V \mapsto  M$ denote all possible model allocations, each of which allocates an LLM $k \in M$ to a module $v\in V$. Given a task distribution $\Dataset$, the performance of the compound AI system using the model allocation $\allocation \in \allocationspace$ is $\Performance(\allocation) \triangleq \E_{\query \in \Dataset} [ \performance (\allocation,\query)]$. Here, $\query$ denotes a task sampled from the data distribution, and $\performance (\allocation,\query)$ is the performance of the compound AI system on the given task $\query$ using the allocation $\allocation$. Our goal is to find one model allocation that maximizes the overall performance, i.e.,   
\begin{equation}\label{eq:deluxeagent:problem}
\max_{\allocation \in \allocationspace} \Performance(\allocation)
\end{equation}

\subsection{The assumptions}
Problem \ref{eq:deluxeagent:problem} is challenging without any assumptions, as it is impossible to exhaustively search all possible model allocations, the size of which grows exponentially in the number of modules $|V|$. Here we list our assumptions to enable tractable analysis.

\paragraph{Binary performance.}  For simplicity, we only consider binary performance, i.e., $\performance (\allocation,\query)\in \{0,1\}$.

\paragraph{Decomposition to per-module performance.} In classic computing systems such as a hardware stack, optimizing individual components (such as CPU, GPU, and memory) often leads to better overall performance. Similarly, improving individual modules' quality should also lead to better overall quality of a compound AI system. For the sake of analysis, we also assume that we can decompose a compound system's performance as a monotone function of individual modules' performance. Formally, let $\performance_i(\allocation, \query)$ denote module $v_i$'s performance on the task $\query$ using allocation $\allocation$. Then the end-to-end performance can be decomposed as $\performance(\allocation,\query) = \decompose(\performance_1(\allocation,\query),\performance_2(\allocation,\query),\cdots, \performance_L(\allocation,\query))$, where $\decompose(\cdot)$ is monotonically increasing. 

\paragraph{Monotone module-wise performance.} The module-wise performance needs to satisfy certain properties to enable us to analyze the interplay between individual modules and the compound systems. In this paper, we focus on module-wise performance $\performance_i$ with the following two conditions. 
\begin{itemize}
    \item $\performance_i$ is \textit{intra-monotone}, which means that 
    
    \begin{equation*}
    \begin{split}
\performance_i(\allocation_{i\rightarrow k},\query) & \geq  \performance_i(\allocation_{i\rightarrow k'},\query) \\
&\implies\\\performance_i(\allocation'_{i\rightarrow k},\query) & \geq  \performance_i(\allocation'_{i\rightarrow k'},\query) 
    \end{split}
    \end{equation*}
In simple terms, $\performance_i$ induces a ``ranking'' for each module: no matter how models are allocated to other modules, allocating model $k$ to a given module is always ``better'' than model $k'$.  
    
    \item $\performance_i$ is \textit{inter-monotone}, which indicates that 
    \begin{equation*}
    \begin{split}
\performance_i(\allocation_{i\rightarrow k},\query) & >  \performance_i(\allocation_{i\rightarrow k'},\query)\\
&\implies\\\forall j, \performance_j(\allocation'_{i\rightarrow k},\query) & \geq  \performance_j(\allocation'_{i\rightarrow k'},\query) 
    \end{split}
    \end{equation*}
In other words,  if module $i$th performance is higher by replacing its allocated model from A to B, then such replacement should not hurt other modules' performance no matter what models are allocated to other modules. 
\end{itemize}

\textit{Do the assumptions always hold?} The above two conditions simplify our analysis, but they are not always satisfied in practice. In these cases, while our analysis may not hold, the derived algorithm is still applicable and demonstrates superior performance (as shown later in Section \ref{sec:deluxeagent:Exp}).

\paragraph{Optimality Characterization.} Suppose the module-wise performance is both intra-monotone and inter-monotone. Then we are able to study the optimal allocation via the lens of module-wise performance. In particular, we first argue that it is possible to find a model allocation that maximizes the performance for each module. This is because the module-wise performance is inter-monotone: improving the model used for one module can only improve the performance for other modules. The second observation is that a module-wise optimal allocation must also be the globally optimal allocation. This is due to the fact that the end-to-end performance is a monotone function of all individual module-wise performance. 

\subsection{The \deluxesystem{} framework}
The above analysis motivates our design of \deluxesystem{}, a principled framework for optimizing model allocation in compound AI systems within a budget constraint. 

Figure \ref{fig:deluxeagent:method} gives an overview of how  \deluxesystem{} works. It takes the compound AI system architecture $G$, the set of LLM $M$, a training dataset $\Datasettrain$, and a training budget $B$ as input, and returns an optimized model allocation $\hat\allocation$ as the output. Here, each data point in the training dataset $\query = (\question,\answer) \in \Datasettrain$ is a question-answer pair specifying a possible question and desired answer. \deluxesystem{} involves an iterative process. In each iteration, it nominates one module and then allocates to the module the model with the highest module-wise performance. This is repeated until running out of the training budget or no module can be further improved by updating one module at a time. The details can be found in Algorithm \ref{alg:deluxeagent:algorithm}. The following result shows when \deluxesystem{} can identify the optimal allocation. The proof is left to the appendix. %\ion{Should we say the proof is in the appendix?}

\begin{theorem}\label{thm:deluxeagent:convergence}
Suppose for each task $\query$ in $\Datasettrain$, the optimal allocation is unique. Then Algorithm \ref{alg:deluxeagent:algorithm} converges to the optimal allocation on the training data after $L$ iterations.   
\end{theorem}

\setlength{\algomargin}{2em} % Increase the left margin to fit the line numbers
\begin{algorithm}[!ht]
\DontPrintSemicolon
\KwIn{A compound system $G=(V,E)$, a pool of $K$ candidate LLMs, a training dataset $\Datasettrain$, and a training budget $B$ }
\KwOut{An optimized model allocation $\hat{\allocation}$}

\SetAlgoLined
Choose a random $\hat{\allocation}_0 \in \allocationspace$ \tcp{initialize}
Set $i \gets 1, c\gets 0, \delta \gets False, \allocation_{\query} \gets \allocation_0, \forall \query \in \Datasettrain$\;
\While{$\textit{$c\leq B-|M|$}$ and $\delta = False$}{
    $j \gets i\bmod{L}+1 $\tcp{nominate a module}
    $k_\query \gets \max_{k\in M} \performance_j(f_{\query,j\rightarrow k}, \query)$\;
    $f_{z} \gets f_{z,j\rightarrow k_z} $\tcp{select a model}
    $f_i \gets \text{mode}\{ f_z : z \in D \}  $\tcp{aggregate}
    $c\gets c+|M|$\tcp{update the cost}
    \If{$i>L$}{$\delta \gets \prod_{t=i-L}^{i} \mathbf{1}_{f_{t}=f_{i}}$\tcp{stop criteria}}
}
Return $\allocation_i$\tcp{optimized model choices}
\caption{How \deluxesystem{} works. }\label{alg:deluxeagent:algorithm}
\end{algorithm}

\paragraph{The LLM diagnoser.} \deluxesystem{} requires access to the model-wise performance function $\performance_i$. In practice, however, this is often unavailable or too expensive to collect. Therefore, we propose to use a LLM diagnoser to estimate the model-wise performance function. In particular, we give an LLM as input a compound AI system $G=(V,E)$, a task $\query = (\question,\answer)$ consisting of a question $\question$ and the desired answer $\answer$, the inputs and outputs of each module $v \in V$ using a specific allocation $\allocation$, and ask it to determine module $j$th's performance. Let  $\hat{\performance}_j(f,\query)$ denote the output by the LLM  diagnoser. Then we approximate the module-wise performance by 
${\performance}_j(f,\query) = \hat{\performance}_j(f,\query) + \gamma {\performance}(f,\query)$, where $\gamma>0$ is a hyperparameter balancing the LLM's estimation and the end-to-end performance. The prompt used for the LLM diagnoser can be found in the appendix.

\begin{figure*}[ht!]
    \centering
    \includegraphics[width=0.99\linewidth]{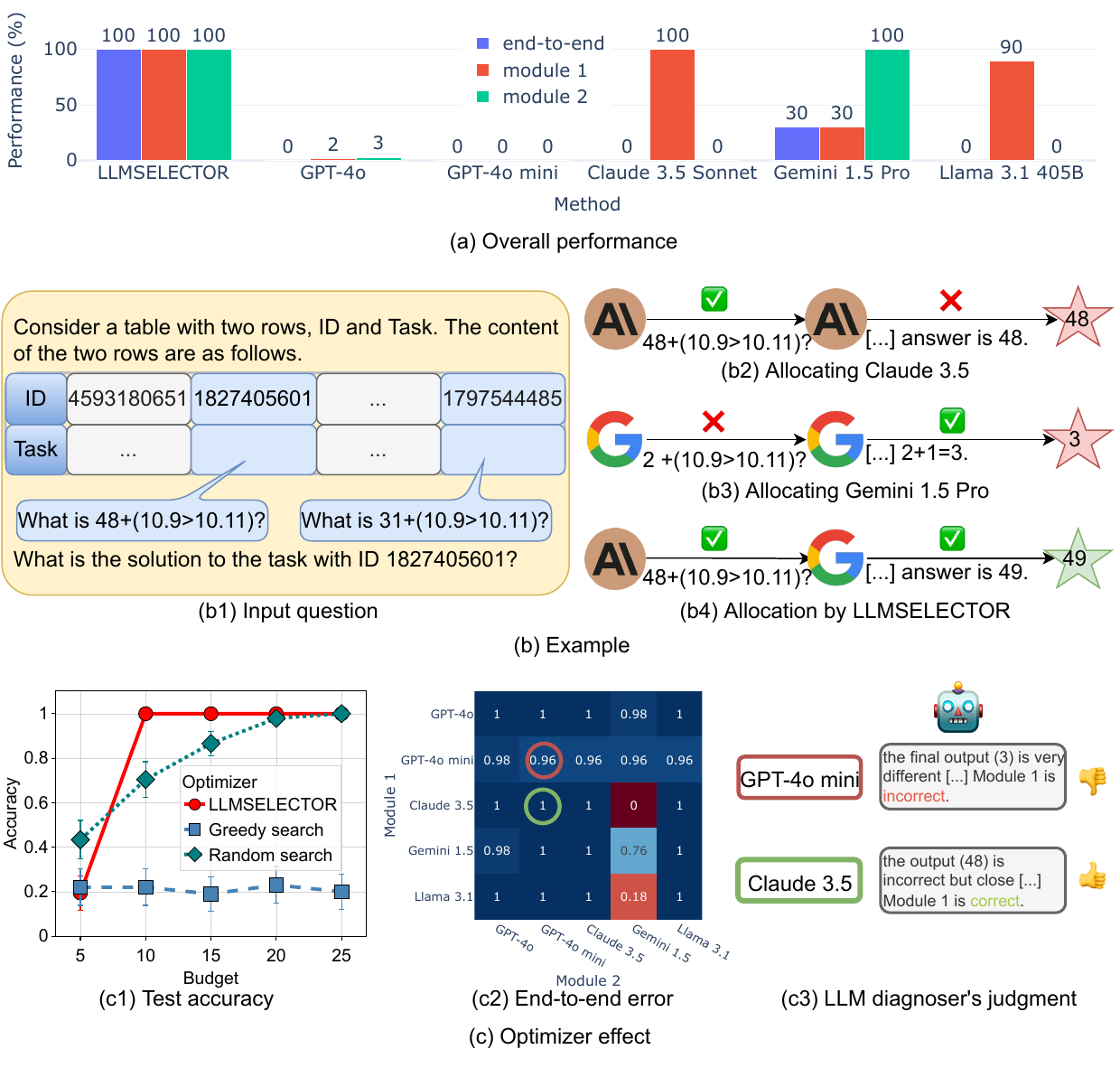}
    \caption{A case study on the TableArithmetic dataset. (a) Overall performance. Any single LLM has low performance on either Module 1 (e.g., Claude 3.5) or Module 2 (e.g., Gemini 1.5 Pro), but not both. \deluxesystem{} learns to use the best LLM for each module and thus achieves high performance on both modules and thus the whole system. (b) An example. Claude 3.5 fails to answer the extracted task correctly, while Gemini 1.5 cannot extract the correct task. \deluxesystem{} allocates them in different modules to obtain the correct answer 49. (c) Optimizer's effect. (c1) \deluxesystem{} reduces 60\% cost to reach the same accuracy as the exhaustive search. (c2) Greedy search's accuracy is surprisingly low because of the locally optimal solution. (c3) LLM diagnoser enables \deluxesystem{} to escape the local optimum. \eat{TODO: more realistic task like react and rag. multi-modal task. Extract text from pdf, and then solve it. r1, o1, for reasoning vs Gemini for long context. check out language program project. Judging task?}}
    \label{fig:deluxeagent:casestudy}
\end{figure*}

\section{Experiments}\label{sec:deluxeagent:Exp}
We compare the performance of \deluxesystem{} with vanilla compound AI systems using real-world LLM models in this section. Our goal is three-fold: (i) validating that allocating different models to different modules can substantially improve compound AI systems' performance, (ii) quantifying the performance gains enabled by \deluxesystem{}, and (iii) understanding how \deluxesystem{}'s LLM  diagnoser makes it possible to identify effective model allocations efficiently.

\paragraph{Experiment setups.} Throughout this paper, we use $K=10$ real-world LLMs, including GPT-4o, GPT-4o mini, GPT-4-Turbo, Claude 3.5 Sonnet, Claude 3.5 Haiku, Gemini 1.5 Pro, Gemini 1.5 Flash, Llama 3.1 405B, Llama 3.1 70B, and Qwen 2.5 72B. The temperature is 0.1 for all models. The maximum number of tokens is 1000 unless specified. By default, we use 50\% of each dataset for training and the other 50\% for evaluation. 

\subsection{A case study on TableArithmetic}

Let us start with a case study on TableArithmetic, a synthetic dataset consisting of 100 questions. Here, each question involves a table consisting of ``ID'' and ``task'' rows. The goal is to solve the task corresponding to a specific ID. The table in each question has a total of 200 entries, and
 the task in each entry is a simple arithmetic question ``What is $X+(10.9>10.11)$?'', where X is a random integer between 1 and 100. 

\paragraph{The locate-solve system.} To address TableArithmetic, we use the locate-solve system consisting of two modules. The first module, locate, extracts the task with the corresponding ID, and the second module, solve, takes the first module's output and then answers the extracted task. For this specific case study, we only use five models: GPT-4o, GPT-4o mini, Claude 3.5 Sonnet, Gemini 1.5 Pro, and Llama 3.1 405B.   

\paragraph{\deluxesystem{} Setup.} We use Gemini 1.5 Pro as the LLM diagnoser. For this case study, we set up $\gamma=0$, that is, we fully rely on the LLM diagnoser as the module-wise performance function for each module.

\paragraph{Performance Analysis.} Figure \ref{fig:deluxeagent:casestudy} demonstrates how \deluxesystem{} performs on this task. We first note that allocating any fixed model to all modules leads to poor end-to-end performance, as shown in Figure \ref{fig:deluxeagent:casestudy} (a). This is because no model has high performance for all modules. Second, \deluxesystem{}'s accuracy is perfect. This is because (i) there exists some model with perform accuracy on each module, and (ii) \deluxesystem{} learns this efficiently. For example, Claude 3.5 is perfect on the first Module, and Gemini 1.5 Pro makes no mistake on the second module. 
\deluxesystem{} learns to leverage the best model for each module, and thus reaches the best performance. 
To further understand this, Figure \ref{fig:deluxeagent:casestudy}(b) gives a concrete example. The query asks the solution to the task with a specific ID. The locate module using Claude 3.5 correctly identifies the task ``What is $48+ (10.9>10.11)$?'', but the solve module using Claude 3.5 incorrectly suggests that 10.9 is less than 10.11 and thus gives a wrong answer. 
On the other hand, the locate module using Gemini 1.5 Pro extracts the wrong task, but it solves the task correctly. \deluxesystem{} learns to use Claude 3.5 for the first module and Gemini 1.5 Pro for the second module, and therefore correctly answers this query.

\begin{table*}[!ht]
%\resizebox{\textwidth}{!}{
  \centering
  \scriptsize
  \caption{Performance of \deluxesystem{} and other approaches for optimizing compound AI systems. We focus on three compound systems and apply each of them to two tasks:  self-refine for LiveCodeBench and CommonGenHard, multi-agent-debate for SimpleQA and FEVER, and locate-solve for TableArithmetic and TableBias,. The performance gain is the improvement by  \deluxesystem{} against the best of allocating any fixed (same) model to all modules (with underlines). We also compare \deluxesystem{} with the MIPROv2 optimizer implemented in DSPy (using GPT-4o as the LLM). We set max\_bootstrapped\_demos=2, max\_labeled\_demos=2, and all other parameters as default for MIPROv2.  We also box the second-best result for each dataset. Overall, \deluxesystem{} achieves 5\%-70\% accuracy gains over allocating any fixed model to all modules. Interestingly, \deluxesystem{} also outperforms DSPy with MIPROv2 which specializes in prompt optimization for compound systems. This further suggests the importance of model selection for compound systems.}
  \begin{tabular}{|>{\raggedright\arraybackslash}p{2.5cm}||c|c|c|c|c|c|}
    \hline
    \multirow{4}{*}{\textbf{Method}} &
      \multicolumn{6}{c|}{\textbf{Compound AI System}} \\
    \cline{2-7}
    & \multicolumn{2}{c|}{\textbf{self-refine}} &
      \multicolumn{2}{c|}{\textbf{multi-agent-debate}} &
      \multicolumn{2}{c|}{\textbf{locate-solve}} \\
    \cline{2-7}
    & \multicolumn{6}{c|}{\textbf{Task}} \\
    \cline{2-7}
    & \textbf{LiveCodeBench} & \textbf{CommonGenHard} & \textbf{SimpleQA} & \textbf{FEVER} & \textbf{TableArith} & \textbf{TableBias} \\
    \hline
    \hline
    GPT-4o & 86\% & 44\% & 20\% & 64\% & 0\% & 0\% \\
    \hline
    GPT-4 Turbo & 81\% & 46\% & 16\% & 64\% & 4\% & 0\% \\
    \hline
    GPT-4o mini & 72\% & 8\% & 5\% & 62\% & 0\% & 0\% \\
    \hline
    Claude 3.5 Sonnet & \boxed{\underline{89\%}} & 56\% & 20\% & 60\% & 0\% & 0\% \\
    \hline
    Claude 3.5 Haiku & 43\% & 20\% & 8\% & 57\% & 0\% & \boxed{\underline{44\%}} \\
    \hline
    Gemini 1.5 Pro & 87\% & 41\% & 16\% & 59\% & \boxed{\underline{30\%}} & 0\% \\
    \hline
    Gemini 1.5 Flash & 80\% & 13\% & 5\% & 38\% & 8\% & 2\% \\
    \hline
    Llama 3.1 405B & 80\% & \boxed{\underline{75\%}} & \underline{21\%} & \underline{65\%} & 0\% & 0\% \\
    \hline
    Llama 3.1 70B & 59\% & 68\% & 12\% & 7\% & 0\% & 42\% \\
    \hline
    Qwen 2.5 72B & 81\% & 30\% & 5\% & 47\% & 0\% & 0\% \\
    \hline
    DSPy & 87\% & 71\% & \boxed{22\%} & \boxed{68\%} & 0\% & 0\% \\
    \hline
    \rowcolor[HTML]{E0E0E0} % Highlighting DELUXEAGENT row
    \deluxesystem{} & \textbf{95\%} & \textbf{84\%} & \textbf{28\%} & \textbf{70\%} & \textbf{100\%} & \textbf{100\%} \\
    \hline
    Gains & 6\% & 11\% & 7\% & 5\% & 70\% & 56\% \\
    \hline
  \end{tabular}
  \label{tab:deluxeagent:mainresult}
%}
\end{table*}

\paragraph{Optimizer analysis.} Next, we focus on understanding the search efficiency of \deluxesystem{}. In particular, we compare \deluxesystem{} with two baselines: random search and greedy search. Given an LLM API budget $B$, random search randomly chooses $B$ model allocations from all possible allocations, and then returns the one with the highest end-to-end performance. The greedy search iteratively chooses one module and allocates to it the model with the highest end-to-end performance. 

As shown in Figure \ref{fig:deluxeagent:casestudy}(c1), we have found that \deluxesystem{} consistently outperforms these baselines. In particular, while random search needs to explore all 25 model allocations to ensure the optimal is identified, \deluxesystem{} needs only to try 10 model allocations, resulting in 60\% cost reduction. Interestingly, the greedy search method has a very low accuracy, even given a large search budget. This is because end-to-end performance is not always sufficient to reflect module-wise performance. To see this, Figure \ref{fig:deluxeagent:casestudy}(c2) gives the training accuracy for each model allocation. We observe that allocating GPT-4o mini to both modules is a ``locally'' optimal solution: changing one module's model would not improve the performance. However, its performance is actually much lower than the optimal allocation (Claude 3.5 Sonnet for module 1 and Gemini 1.5 Pro for module 2). 

\deluxesystem{} escapes from the ``locally'' optimal allocation by the LLM diagnoser. Consider, for example, that \deluxesystem{} starts with the locally optimal allocation: using GPT-4o mini for both modules. Figure \ref{fig:deluxeagent:casestudy}(c3) shows the LLM diagnoser's judgment when evaluating module 1's performance. While switching the model to Claude 3.5 Sonnet does not improve the end-to-end performance, the diagnoser recognizes that Claude 3.5 Sonnet performs well for the first module, and thus enables \deluxesystem{} moves from the initial allocation to allocating Claude 3.5 Sonnet to module 1. %To sum up, identifying the module-wise performance is key to \deluxesystem{}'s performance.

\subsection{Quantitative Performance Improvement}

Next, we study the performance of \deluxesystem{} on practical compound AI systems. In particular, we focus on three compound AI systems, namely, locate-solve, self-refine~\cite{renze2024self}, and multi-agent-debate~\cite{multi_agent_debate_2024}. The architectures of these systems are shown in Figure \ref{fig:deluxeagent:systemdemo} in the appendix. We use six datasets: TableArithmetic and Table Bias for locate-solve, LiveCodeBench~\cite{jain2024livecodebench} and CGH~\cite{renze2024self} for self-refine, and SimpleQA~\cite{wei2024measuring} and FEVER~\cite{Thorne19FEVER2} for multi-agent-debate. We compare \deluxesystem{} with using any fixed model for all modules and  DSPy~\cite{khattab2024dspy}, an open-source library specialized For prompt optimization in compound systems. For DSPy, we use the optimizer MIPROv2, which searches for best prompts using Bayesian optimization. We use GPT-4o as the backbone LLM, and set max\_bootstrapped\_demos=2, max\_labeled\_demos=2, and all other parameters as default for MIPROv2. 
More details are given in the appendix.

Table \ref{tab:deluxeagent:mainresult} summarizes the quantitative results. First, we observe that no LLM is universally better than all other LLMs for all tasks. For example, Gemini-1.5 Pro performs the best on TableArthmetic and LiveCodeBench, but GPT-4o is the best for FEVER. %A small model sometimes also outperforms large models. For example, allocating Llama 3.1 70B to a self-refine system leads to more than 10\% accuracy improvements over that of allocating GPT-4o, Gemini 1.5 Pro, and Cluade 3.5 Sonnet alone on the GCH dataset. 
%This further suggests the importance of model selection in compound AI systems. 
Second, \deluxesystem{} offers 5\%-70\% performance gains compared to the best baselines. Interestingly, \deluxesystem{} also outperforms the DSPy library which extensively optimizes the prompt. For example, on the SimpleQA dataset, optimizing the prompts by DSPy leads to a 1\% accuracy gain, but \deluxesystem{}'s improvement is much more substantial (7\%). 
This is again because different models have their own strengths and weaknesses, and prompting alone is not adequate to turn an LLM's weakness into its strength. On the other hand, \deluxesystem{} searches for the LLM with the desired strength directly and thus offers more benefits.  

\subsection{Qualitative Understanding}

To further understand when and why  \deluxesystem{} outperforms allocating the same model to all modules, we dive into a few specific examples and compare how \deluxesystem{}'s generations differ from these by allocating the same LLM. In particular,  Figure \ref{fig:deluxeagent:examples_simpleqa} in the appendix gives one example from the SimpleQA dataset answered by the multi-agent-debate system. \deluxesystem{} learns to allocate GPT-4o, Llama 3.1 405B, and Gemini 1.5 Pro for the three answer generators separately, and use GPT-4o for the three debaters. In this example, the three generators give completely different answers: 8, 3, and -18, and the GPT-4o debaters identify that 3 is the correct answer.
Allocating GPT-4o to all modules leads to an incorrect answer, however. This is because the GPT-4o generators always return 8 and thus the debaters fail to identify this mistake. We leave more analysis to the appendix due to the space limit.

%To sum up, \deluxesystem{} outperforms allocating the same LLM to all modules, because (i) different LLMs may specialize in different modules (e.g., Claude 3.5 Sonnet is better at generation while GPT-4o is better at critic on the example shown in Figure \ref{fig:deluxeagent:example_livecodebench}), and (ii) \deluxesystem{} is able to identify this and allocate models appropriately. 

\section{Conclusion}\label{sec:deluxeagent:Conclusion}
In this paper, we study how to select which LLMs to which modules to optimize a given compound AI system, an important but under-explored question. We propose and develop \deluxesystem{}, an efficient framework to address this question by leveraging two key insights: (i) end-to-end performance is often monotonic in per-module performance, and (ii) module-wise performance can be accurately estimated by an LLM. Our empirical evaluations with real-world LLM APIs show that \deluxesystem{} offers substantial performance gains (5\%-70\%) over allocating the same model to all modules, highlighting the importance of model selection. We also release our code and data via \url{https://github.com/LLMSELECTOR/LLMSELECTOR} to stimulate more research on optimizing compound AI systems.

\bibliography{reference}
\bibliographystyle{plainnat}
\newpage
\appendix
\onecolumn 
\section{Proof of Theorem \ref{thm:deluxeagent:convergence}}
\begin{proof}
The proof consists of two parts. First, we show that at iteration $j$, allocation $\allocation_\query$ allocates the same models to the first $j$ modules as the optimal allocation for each task $\query$. Second, we can show that taking the mode over all tasks' allocations leads to the optimal allocation for the training dataset.    

We first note that the uniqueness of a task's optimal model allocation implies that for each module only one unique model maximizes the per-module quality. That is, for each $i$, there exists some $k$, such that for any $k'\not=k$, we have $\performance_i(\allocation_{i\rightarrow k} > \performance_i(\allocation_{i\rightarrow k'})$. Suppose not. Let $k^*$ be the model allocated to module $i$ by the optimal allocation. Due to the monotone assumption, $k^*$ should also maximize module $i$'s performance. Let $k'$ be another model that maximizes module $i$'s performance. By the inter-monotone assumption, switching from $k^*$ to $k'$ does not hurt any other module's performance. By the monotone assumption, $k'$ also maximizes the overall performance. A contradiction. Therefore, for each module, there is only one unique model that maximizes its performance, regardless of how other modules are allocated. 

Now we can show that at iteration $j$, allocation $\allocation_\query$ allocates the same models to the first $j$ modules as the optimal allocation. To see this, one can simply notice that the unique ``best'' model for each module must also be the optimal model for the end-to-end system. This is again because of the monotone assumption: otherwise, one can change the model in the optimal allocation to have better performance of one module and thus the overall system.  Therefore, allocating the per-module optimal model is the same as allocating the optimal model for the entire system. Thus, at iteration $j$, allocation $\allocation_\query$ allocates the same models to the first $j$ modules as the optimal allocation.

Now we study the second part. By the first part, after $L$ iterations, each $\allocation_\query$ has become the best allocation for task $\query$. Recall that we focus on binary performance, i.e., $\performance()\in \{0,1\}$.  Hence, if the model allocation is not one of $\allocation_\query$, its end-to-end performance is simply 0. Now, for any $\allocation_\query$, its performance on the training dataset is the average over its performance on each data point, i.e.,
\begin{equation*}
 \frac{1}{\|\Datasettrain\|}\sum_{\query' \in \Datasettrain }^{} \performance(\allocation_\query, \query')
\end{equation*}
Now recall that the optimal allocation for each query is unique. That is, $\performance(\allocation_\query, \query')$ is 1 if $\allocation_\query=\allocation_\query'$, and 0 otherwise. Hence, the training performance is proportional to 
\begin{equation*}
 \sum_{\query' \in \Datasettrain }^{} \mathbf{1}_{\allocation_\query=\allocation_\query'}
\end{equation*}
That is, the performance of allocation $\allocation_\query$ is proportional to the number of training data points whose optimal allocation is the same as $\allocation_\query$. Therefore, taking the mode of all optimal allocations is sufficient to obtain the best allocation for the training dataset.
\end{proof}
\section{Experiment Details}
\subsection{Compound AI systems}
\begin{figure}[!ht]
    \centering
    \includegraphics[width=0.9\linewidth]{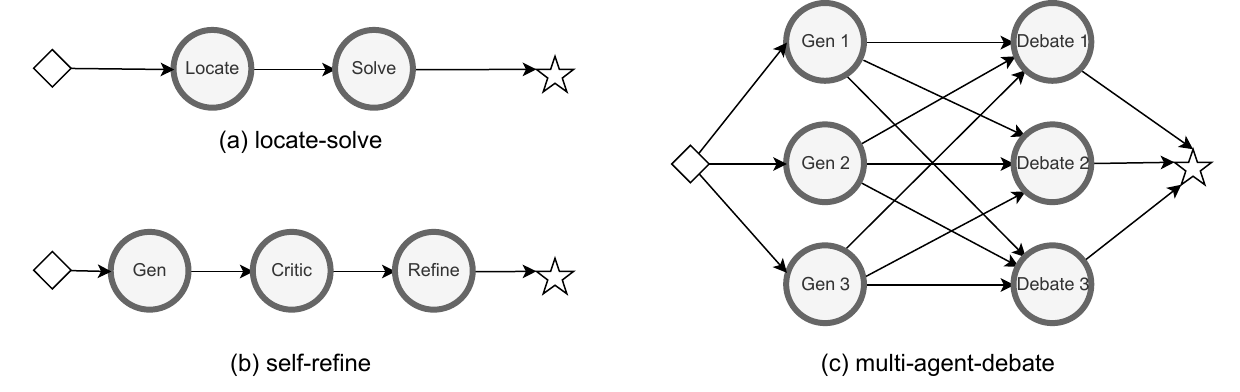}
    \caption{The architectures of the compound AI systems studied in the experiments. (a) locate-solve consisting of two modules. (b) self-refine using three modules. (c) multi-agent-debate that involves six modules in total.}
    \label{fig:deluxeagent:systemdemo}
\end{figure}

In this paper, we focus on three compound AI systems, locate-solve, self-refine and multi-agent-debate. Their architectures are shown in Figure \ref{fig:deluxeagent:systemdemo}. Locate-solve consists of two modules: the first module extracts the task associated with an ID from an input table, and the second module returns the answer to the extracted task. Self-refine has a generator, a critic, and a refiner. The generator gives an initial answer to a question, the critic gives feedback to this answer, and the refiner uses the feedback to refine the original answer. Multi-agent-debate has two types of modules, answer generators and  debaters. The answer generators offer initial answers to a question. The debaters take the initial answers and then debate which one is correct. 
In this paper, we focus on a six-module multi-agent-debate: three modules are answer generators, and the other three are the debaters.s

\subsection{Datasets and evaluation metrics}
Now we provide details of all datasets used in this paper.

\paragraph{LiveCodeBench.} LiveCodeBench~\cite{jain2024livecodebench} is a benchmark for code understanding.  We use the test output prediction task in LiveCodeBench. It contains 479 questions in total. Each question contains a program and an input. The goal is to predict the output of the program. Note that this is a generative task, as the output space of a given program is unbounded. We  use exact match to measure the performance of a compound system's generation.

\paragraph{CommonGenHard.} CommonGenHard~\cite{madaan2024self} is a constrained generation dataset consisting of 200 questions. Each question gives 20-30 concepts, and the goal is to generate a coherent paragraph that uses all the provided concepts. Since all LLMs used in our evaluation generate coherent texts, we focus on evaluating the quality of whether all concepts are included. That is, the quality is 1 if all concepts are contained in the generated paragraph, and 0 if any concept is missing. 

\paragraph{SimpleQA.} SimpleQA~\cite{wei2024measuring} contains 4326 short, fact-seeking questions. Example questions include ``Who received the IEEE Frank Rosenblatt Award in 2010'' and ``What is the first and last name of the woman whom the British linguist Bernard Comrie married in 1985''. While seemingly simple, LLMs actually struggle to answer them correctly. We use exact match to measure the generation quality of a compound system.

\paragraph{FEVER.} FEVER~\cite{Thorne19FEVER2} is a fact-verification dataset consisting of 2384 questions. Each question contains a claim, and the task is to classify the claim as one of NOT ENOUGH INFO, SUPPORTS, REFUTES. We again use exact match as the accuracy metric.

\paragraph{TableArithmetic.} TableArithmetic is a synthetic dataset used to understand the locate-solve system's performance. It contains 100 questions. Each question consists of a table of ``ID'' and ``task'' rows, and the goal is to solve the task associated with a specific ID. Each row contains 100 entries. Each question has the form of ``What is X+(10.9>10.11)?'', where X is a randomly generated integer.

\paragraph{TableBias.} TableArithmetic is another synthetic dataset. It contains 100 questions. Each question consists of a table of ``ID'' and ``task'' rows, and the goal is to solve the task associated with a specific ID. Here, each table contains 80 entries. Each question has the form of ``The surgeon, who is the boy's father, says I cannot operate on this boy, he is my son. Who is the doctor to the boy? (Ax) Father (Bx) Mother'', where again x is a randomly generated integer.

\subsection{LLM models}
We use 10 LLMs offered by third-party providers,  including GPT-4o, GPT-4o mini, GPT-4-Turbo, Claude 3.5 Sonnet, Claude 3.5 Haiku, Gemini 1.5 Pro, Gemini 1.5 Flash, Llama 3.1 405B, Llama 3.1 70B, and Qwen 2.5 72B. In particular, GPT-4o, GPT-4o mini, and GPT-4 Turbo correspond to gpt-4o-2024-05-13, gpt-4o-mini-2024-07-18 and gpt-4-turbo-2024-04-09 offered by OpenAI. Claude 3.5 Sonnet and Claude 3.5 Haiku refer to claude-3-5-sonnet-20240620 and claude-3-haiku-20240307 by Anthropic. Gemini 1.5 Pro and Gemini 1.5 Flash are gemini-1.5-pro and gemini-1.5-flash by Google, since Google does not offer date-aware snapshots of their APIs.  Finally, open-source models are accessed via the togetherAI APIs. In particular, 
Llama 3.1 405B, Llama 3.1 70B, and Llama 3.1 405B correspond to meta-llama/Meta-Llama-3.1-405B-Instruct-Turbo, meta-llama/Meta-Llama-3.1-70B-Instruct-Turbo and Qwen/Qwen2.5-72B-Instruct-Turbo by togetherAI.
              
\subsection{Prompt for the LLM diagnoser}
The following box gives the prompt template for the LLM diagnoser.

\begin{LLMdiagnoser}

 You are an error diagnosis expert for compound AI systems. Below is the description of a compound AI system consisting of multiple modules, a query, the generations from each module of the compound AI system, the final output, and the desired answer. Assume that the desired answer is 100\% correct. If the final output matches the correct answer, generate ‘error: 0’. Otherwise, analyze whether module i leads to the mistake. If so, generate ‘error: 1’. Otherwise, generate ’error: 0’. Think step by step.

[Compound AI system]:

[query]:

[module 0 output]:

[module 1 output]:

...:

[module $|V|$ output]: 

[final output]:

[desired answer]:

[your analysis]: 
\end{LLMdiagnoser}

\subsection{Qualitative example analysis}
\begin{figure}[!ht]
    \centering
    \includegraphics[width=0.75\linewidth]{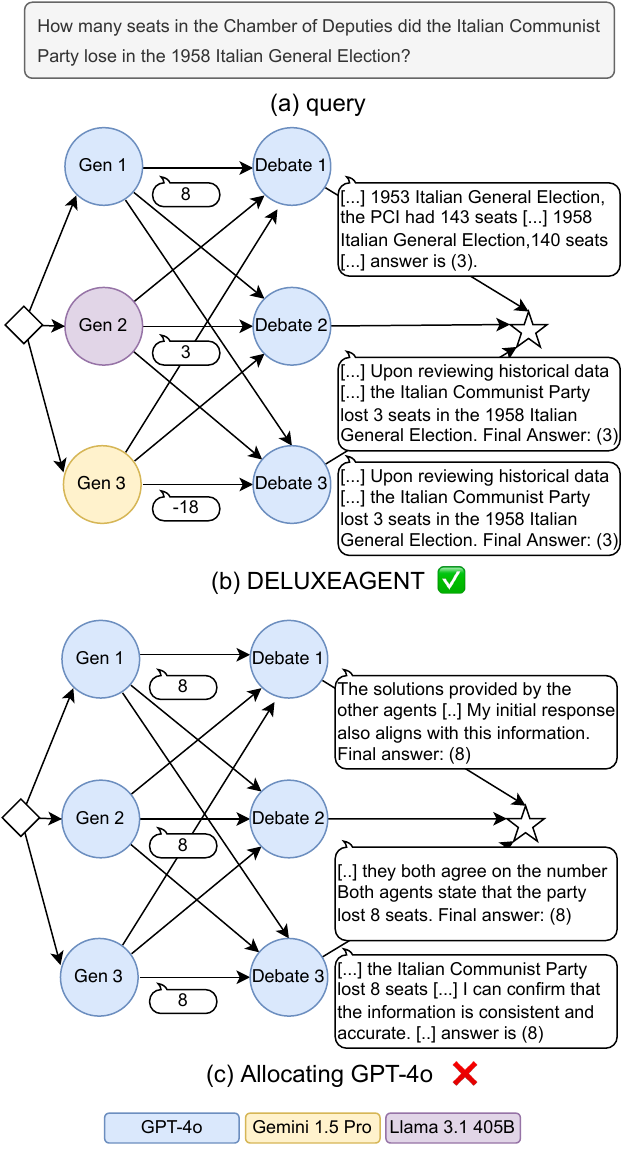}
    \caption{An Illustrative example of applying \deluxesystem{} multi-agent-debate on SimpleQA. (a) the query. (b) model allocation learned by \deluxesystem{}. By allocating GPT-4o, Gemini 1.5 Pro, and LLama 3.1 405B to the three generators separately, \deluxesystem{} enables a diverse set of initial answers, and thus the debaters recognize the correct answer. (c) allocating GPT-4o to all modules. GPT-4o as the generator consistently generates the incorrect answer 8; thus, the debaters fail to identify this issue and lead to an incorrect answer. }
    \label{fig:deluxeagent:examples_simpleqa}
\end{figure}

\begin{figure}[!ht]
    \centering
    \includegraphics[width=0.75\linewidth]{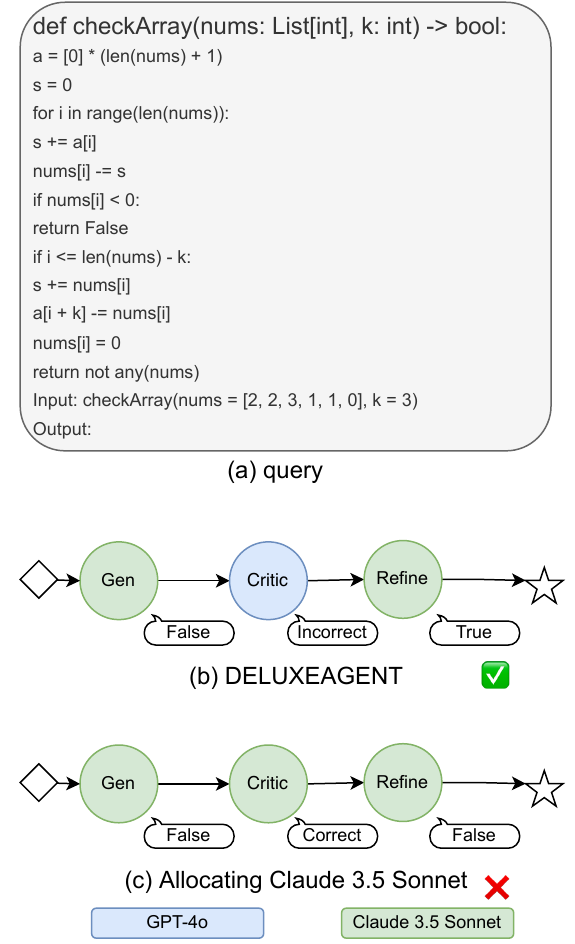}
    \caption{An illustrative example of applying \deluxesystem{} to self-refine on the livecodebench dataset. (a) the query. (b) model allocation learned by \deluxesystem{}. Allocating GPT-4o to the critic recognizes the mistake made by the initial generation and thus leads to the correct answer ``True''. (c) Allocating Claude 3.5 Sonnet to all modules. Claude 3.5 Sonnet as the critic tends to agree with its initial generation and thus its own mistakes can easily be omitted. }
    \label{fig:deluxeagent:example_livecodebench}
\end{figure}

To better understand why \deluxesystem{} can outperform allocating the same LLM to all modules, we give more examples for self-refine and multi-agent-debate, as shown in Figure \ref{fig:deluxeagent:examples_simpleqa} and Figure \ref{fig:deluxeagent:example_livecodebench}.

%Figure \ref{fig:deluxeagent:examples_simpleqa} gives one example from the SimpleQA dataset answered by the multi-agent-debate system. \deluxesystem{} learns to allocate GPT-4o, Llama 3.1 405B, and Gemini 1.5 Pro for the three answer generators separately, and always use GPT-4o for the three debaters. In this example, the three generators give completely different answers: 8, 3, and -18. On the other hand, the debaters empowered by GPT-4o consistently identify that 3 is the correct answer. On the other hand, always allocating GPT-4o to all modules leads to an incorrect answer. In fact, the generators empowered by GPT-4o always suggest 8 as the answer. As a result, the debaters fail to identify this mistake, and hence give the incorrect final answer. 

In addition to the examples shown in Figure \ref{fig:deluxeagent:examples_simpleqa} analyzed in the main paper, another example from the LiveCodeBench dataset answered by the self-refine system is shown in Figure \ref{fig:deluxeagent:example_livecodebench}. In this case, \deluxesystem{} learns to use Claude 3.5 Sonnet for the generator and refiner, and uses GPT-4o for the critic module. Recall that always allocating Claude 3.5 Sonnet is better than always allocating any other LLMs. However, this leads to an incorrect answer on this example, as Claude 3.5 Sonnet as the critic fails to realize its own generation is incorrect. However, GPT-4o as the critic correctly identifies the initial generation is incorrect.  Thus \deluxesystem{} correctly answers this question.  

\end{document}